\documentclass[11pt,a4paper]{article}
\usepackage[hyperref]{naaclhlt2018}
\usepackage{times}
\usepackage{latexsym}
\usepackage{url}
\usepackage{amsmath}
\usepackage{hhline}
\usepackage[T1]{fontenc}
\usepackage{graphicx}
\usepackage{eurosym}
\usepackage{tabularx}
\graphicspath{ {images/} }
\usepackage{listings}
\usepackage{stfloats}
\usepackage{csquotes}
\usepackage{array}
\usepackage[super]{nth}
\usepackage[inline,shortlabels]{enumitem}
\usepackage{subfig}
\usepackage{amsmath}
\usepackage{multirow}
\usepackage{svg}
\usepackage{float}
\usepackage[scientific-notation=true]{siunitx}
\usepackage{marginnote}
\usepackage[textsize=tiny]{todonotes}
\usepackage{mdframed}
\aclfinalcopy 
 
\title{NTUA-SLP at SemEval-2018 Task 1: Predicting Affective Content in Tweets with Deep Attentive RNNs and Transfer Learning}

\author{ 
	Christos Baziotis$^{1,3}$,  Nikos Athanasiou$^1$,  Alexandra Chronopoulou$^1$, \\ 
	{\bf Athanasia Kolovou$^{1,2}$, Georgios Paraskevopoulos$^{1,4}$, Nikolaos Ellinas$^1$} \\
	{\bf Shrikanth Narayanan$^{4,5}$,  Alexandros Potamianos$^{1,4,5}$} \\\\
	$^1$School of ECE, National Technical University of Athens, Athens, Greece \\
	$^2$ Department of Informatics, University of Athens, Athens, Greece \\
	$^3$ Department of Informatics, Athens University of Economics and Business, Athens, Greece \\
	$^4$ Behavioral Signal Technologies, Los Angeles, CA\\    
	$^5$ Signal Analysis and Interpretation Laboratory (SAIL), USC, Los Angeles, USA\\       
	{\tt cbaziotis@mail.ntua.gr, el12074@central.ntua.gr} \\
	{\tt el12068@central.ntua.gr, akolovou@di.uoa.gr} \\
	{\tt geopar@central.ntua.gr, nellinas@central.ntua.gr}\\ 	  	      		
    {\tt shri@sipi.usc.edu, potam@central.ntua.gr}
}

\date{2018}

\begin{document}

\maketitle

\begin{abstract}
	In this paper we present deep-learning models that submitted to the SemEval-2018 Task~1 competition: \enquote{Affect in Tweets}. We participated in all subtasks for English tweets. 
    We propose a Bi-LSTM architecture equipped with a multi-layer self attention mechanism. The attention mechanism improves the model performance and allows us to identify salient words in tweets, as well as gain insight into the models making them more interpretable.
    Our model utilizes a set of word2vec word embeddings trained on a large collection of 550 million Twitter messages, augmented by a set of word affective features.
    Due to the limited amount of task-specific training data, we opted for a transfer learning approach by pretraining the Bi-LSTMs on the dataset of Semeval 2017, Task 4A.
	The proposed approach ranked \nth{1} in Subtask E \enquote{Multi-Label Emotion Classification},  \nth{2} in Subtask A \enquote{Emotion Intensity Regression} and achieved competitive results in other subtasks.

\end{abstract}

\section{Introduction}
Social media content has dominated online communication, enriching and changing language with new syntactic and semantic constructs that allow users to express facts, opinions and emotions in short amount of text. The analysis of such content has received great attention in NLP research due to the wide availability of data and the interesting language novelties. Specifically the study of affective content in Twitter has resulted in a variety of novel applications, such as tracking product perception~\cite{chamlertwat2012discovering}, public opinion detection about political tendencies~\cite{pla2014political,tumasjan2010predicting}, stock market monitoring~\cite{si2013exploiting,bollen2011twitter} etc. The wide usage of figurative language, such as emojis and special language forms like abbreviations, hashtags, slang and other social media markers, which do not align with the conventional language structure, make natural language processing in Twitter even more challenging. 


\begin{figure}[!t]
\begin{mdframed}
  \captionsetup{farskip=0pt} 
  \subfloat{\includegraphics[scale=0.8,page=2]{intr}\label{fig:intro_att_1}}
  \hfill
  \vskip 10pt
  \subfloat{\includegraphics[scale=0.8,page=1]{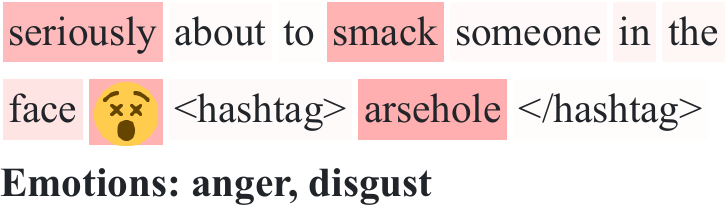}\label{fig:intro_att_2}}
\end{mdframed}

\caption{Attention heat-map visualization. 
The color intensity corresponds to the weight given to each word by the self-attention mechanism.}
\label{fig:intro-att}
\end{figure}
In the past, sentiment analysis was tackled by extracting hand-crafted features or features from sentiment lexicons~\cite{nielsen2011new,mohammad2010emotions,mohammad2013crowdsourcing,go2009twitter} that were fed to classifiers such as Naive Bayes or Support Vector Machines (SVM)~\cite{bollen2011modeling,mohammad2013a,kiritchenko2014}.
The downside of such approaches is that they require extensive feature engineering from experts and thus they cannot keep up with rapid language evolution~\cite{mudinas2012combining}, especially in social media/micro-blogging context.
However, recent advances in artificial neural networks for text classification have shown to outperform conventional approaches~\cite{deriu2016,rouvier2016,rosenthal2017semeval}. This can be attributed to their ability to learn features directly from data and also utilize hand-crafted features where needed. Most of aforementioned works focus on sentiment analysis, but similar approaches have been applied to emotion detection ~\cite{canales2014emotion} leading to similar conclusions. SemEval 2018 Task 1: \enquote{Affect in Tweets}~\cite{SemEval2018Task1} focuses on exploring emotional content of tweets for both classification and regression tasks concerning the four basic emotions (joy, sadness, anger, fear) and the presence of more fine-grained emotions such as disgust or optimism.

\begin{figure}[t]
	\captionsetup{farskip=0pt} 
	\centerline{\includegraphics[width=1.0\columnwidth]{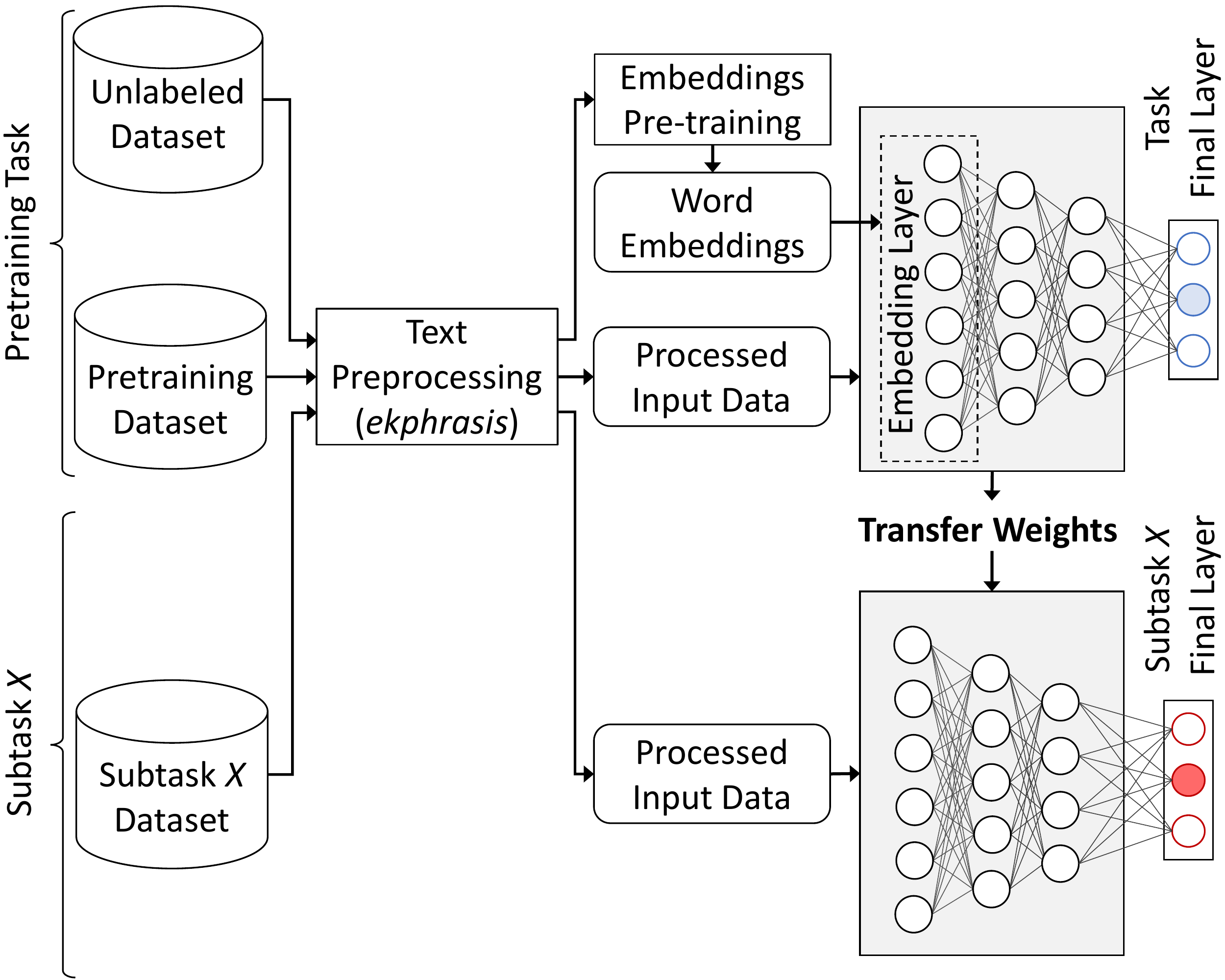}}
	\caption{High-level overview of our approach}
	\label{fig:over}
\end{figure}

In this paper, we present a deep-learning system that competed in SemEval 2018 Task 1: \enquote{Affect in Tweets}. We explore a transfer learning approach to compensate for limited training data that uses the sentiment analysis dataset of Semeval Task 4A~\cite{SemEval:2017:task4} for pretraining a model and then further fine-tune it on data for each subtask. Our model operates at the word-level and uses a Bidirectional LSTM equipped with a deep self-attention mechanism~\cite{pavlopoulos2017deep}. Moreover, to help interpret the inner workings of our model, we provide visualizations of tweets with annotations of the salient tokens as predicted by the attention layer.
\section{Overview} \label{sec:over}
Figure~\ref{fig:over} provides a high-level overview of our approach, which consists of three main steps: 
\begin{enumerate*}[(1)]
	\item the \emph{word embeddings pretraining}, where we train word2vec and affective word embeddings on our unlabeled Twitter dataset,
	\item the \emph{transfer learning} step, where we pretrain a deep-learning model on a sentiment analysis task,
	\item the \emph{fine-tuning} step, where we fine-tune the pretrained model on each subtask.
\end{enumerate*}

\setlist[description]{font=\normalfont\itshape}

\noindent\textbf{Task definitions}.
Given a tweet we are asked to:

\noindent\textit{Subtask EI-reg:} determine the intensity of a certain emotion (\textit{joy, fear, sadness, anger}), as a real-valued number between in the $[0,1]$ interval.

\noindent\textit{Subtask EI-oc:} classify its intensity towards a certain emotion (\textit{joy, fear, sadness, anger}) across a 4-point scale.

\noindent\textit{Subtask V-oc:} classify its valence intensity (i.e sentiment intensity) across a 7-point scale $[-3,3]$.

\noindent\textit{Subtask V-reg:} determine its valence intensity as a real-valued number between in the $[0,1]$ interval.

\noindent\textit{Subtask E-c:} determine the existence of none, one or more out of eleven emotions: \textit{anger, anticipation, disgust, fear, joy, love, optimism, pessimism, sadness, surprise, trust}.


\subsection{Data}
\noindent\textbf{Unlabeled Dataset}. We collected a big dataset of 550 million English tweets, from April 2014 to June 2017. This dataset is used for (1) calculating word statistics needed in our text preprocessing pipeline (Section \ref{sec:prep}) and (2) training word2vec and affective word embeddings (Section \ref{sec:embeddings}).

\noindent\textbf{Pretraining Dataset}.
For transfer learning, we utilized the dataset of Semeval-2017 Task4A. The dataset consists of $61,854$ tweets with $\{positive, neutral, negative\}$ sentiment (valence) annotations. To our knowledge, this is the largest Twitter dataset with affective annotations.

\subsection{Word Embeddings}
\label{sec:embeddings}
Word embeddings are dense vector representations of words~\cite{collobert2008, mikolov2013}, capturing their semantic and syntactic information. To this end, we train \textit{word2vec} word embeddings, to which we add 10 affective dimensions. We use our pretrained embeddings, to initialize the first layer (embedding layer) of our neural networks.

\noindent\textbf{Word2vec Embeddings}.
We leverage our unlabeled dataset to train Twitter-specific word embeddings. 
We use the \textit{word2vec} ~\cite{mikolov2013} algorithm, with the skip-gram model, negative sampling of 5 and minimum word count of 20, utilizing Gensim's~\cite{rehurek_lrec} implementation. The resulting vocabulary contains $800,000$ words.

\noindent\textbf{Affective Embeddings}.
Starting from small manually annotated lexica, continuous norms (within the $[-1, 1]$ interval) for new words are estimated using semantic similarity and a linear model along ten affect-related dimensions, namely: valence, dominance, arousal, pleasantness, anger, sadness, fear, disgust, concreteness, familiarity.
The method of generating word level norms is detailed in  \cite{malandrakis2013distributional} and relies on the assumption that given a similarity metric between two words, one may derive the similarity between their affective ratings. This approach uses a set of $N$ words with known affective ratings (seed words), as a starting point. Concretely, we calculate the affective rating of a word $w$ as follows:

\begin{equation}
\hat{\upsilon}(w) = \alpha_0+\sum_{i=1}^{N}\alpha_i\upsilon(t_i)S(t_i,w),\label{afmod}
\end{equation}
\noindent where $t_{1}...t_{N}$ are the seed words, $\upsilon(t_i)$ is the affective rating for seed word $t_i$, $\alpha_i$ is a trainable weight corresponding to seed $t_i$ and $S(\dot)$ stands for the semantic similarity metric between $t_i$ and $w$. 
The seed words $t_{i}$ are selected separately for each dimension, from the words available in
the original manual annotations (see \ref{annotations}). The $S(\dot)$ metric is estimated as shown in \cite{palogiannidi2015valence} using word-level contextual feature vectors and adopting a scheme based on mutual information for feature weighting.

\noindent\textbf{Manually annotated norms}. \label{annotations}
To generate affective norms, we need to start from some manual annotations, so we use ten dimensions from four sources. From the Affective Norms for English Words \cite{anew} we use norms for valence, arousal and dominance. From the MRC Psycholinguistic database \cite{coltheart1981mrc}, we use 
norms for concreteness and familiarity. From the Paivio norms \cite{clark2004extensions} we use
 norms for pleasantness. Finally from \cite{stevenson2007characterization} we use norms for anger, sadness, fear and disgust.

{\setlength\extrarowheight{0.15em}
	\begin{table*}[!hb]
		\small
		\begin{tabularx}{\linewidth}{ |c|X| }
			\hline
			original  & The *new* season of \#TwinPeaks is coming on May 21, 2017. CANT WAIT \textbackslash o/ !!! \#tvseries \#davidlynch :D                                                                                         \\ 
			\hline
			processed & the new <emphasis> season of <hashtag> twin peaks </hashtag> is coming on <date> . cant <allcaps> wait <allcaps> <happy> ! <repeated> <hashtag> tv series </hashtag> <hashtag> david lynch </hashtag> <laugh> 
			\\ 
			\hline
		\end{tabularx}
		\caption{Example of our text processor}
		\label{table:textpp} 
	\end{table*}
}

\subsection{Preprocessing\footnote{Significant portions of the systems submitted to SemEval 2018 in Tasks 1, 2 and 3, by the NTUA-SLP team are shared, specifically the preprocessing and portions of the DNN architecture. Their description is repeated here for completeness.}}\label{sec:prep}
We utilized the \textit{ekphrasis}\footnote{\url{github.com/cbaziotis/ekphrasis}}~\cite{baziotis2017datastories} tool as a tweet preprocessor. The preprocessing steps included in ekphrasis are: Twitter-specific tokenization, spell correction, word normalization, word segmentation (for splitting hashtags) and word annotation.

\noindent\textbf{Tokenization}. 
Tokenization is the first fundamental preprocessing step and since it is the basis for the other steps, it immediately affects the quality of the features learned by the network. 
Tokenization on Twitter is challenging, since there is large variation in the vocabulary and the expressions which are used. There are certain expressions which are better kept as one token (e.g. anti-american) and others that should be split into separate tokens. 
Ekphrasis recognizes Twitter markup, emoticons, emojis, dates (e.g. 07/11/2011, April 23rd), times (e.g. 4:30pm, 11:00 am), currencies (e.g. \$10, 25mil, 50\euro), acronyms, censored words (e.g. s**t), words with emphasis (e.g. *very*) and more using an extensive list of regular expressions.

\noindent\textbf{Normalization}. 
After tokenization, we apply a series of modifications on the extracted tokens, such as spell correction, word normalization and segmentation.
Specifically for word normalization we use lowercase words, normalize URLs, emails, numbers, dates, times and user handles (@user). This helps reducing the vocabulary size without losing information.
For spell correction \cite{jurafsky2000} and word segmentation \cite{segaran2009a} we use the Viterbi algorithm. The prior probabilities are obtained from word statistics from the unlabeled dataset. 

The benefits of the aforementioned procedure are the reduction of the vocabulary size, without removing any words, and the preservation of information that is usually lost during tokenization.
Table~\ref{table:textpp} shows an example text snippet and the resulting preprocessed tokens.

\subsection{Neural Transfer Learning for NLP}
Transfer learning aims to make use of the knowledge from a source domain, to improve the performance of a model in a different, but related, target domain. It has been applied with great success in computer vision (CV)~\cite{DBLP:journals/corr/RazavianASC14,DBLP:journals/corr/LongSD14}. Deep neural networks in CV are rarely trained from scratch and instead are initialized with pretrained models. Notable examples include face recognition \cite{DeepFaceTaigman} and visual QA ~\cite{Agrawal:2017:VVQ:3088990.3089103}, where image features trained on ImageNet ~\cite{ImageNet} and word embeddings estimated on large corpora via unsupervised training are combined. Although model transfer has seen widespread success in computer vision, transfer learning beyond pretrained word vectors is less pervasive in NLP.

In our system, we explore the approach of pretraining a network in a sentiment analysis task in Twitter and use it to initialize the weights of the models of each subtask. We chose the dataset of Semeval 2017 Task4A (SA2017)~\cite{SemEval:2017:task4}, which is a semantically similar dataset to the emotion datasets of this task. By pretraining on a dataset in a similar domain, it is more likely that the source and target dataset will have similar distributions.

To build our pretrained model, we initialize the weights of the embedding layer with the word2vec Twitter embeddings and train a bidirectional LSTM (BiLSTM) with a deep self-attention mechanism~\cite{pavlopoulos2017deep} on SA2017, similar to \cite{baziotis2017datastories}. Afterwards, we utilize the encoding part of the network, which is the BiLSTM and the attention layer, throwing away the last layer. This pretrained model is used for all subtasks, with the addition of a subtask-specific final layer for classification/regression.
 
\subsection{Recurrent Neural Networks}
We model the Twitter messages using Recurrent Neural Networks (RNN). RNNs process their inputs sequentially, performing the same operation, $ h_t=f_W(x_t, h_{t-1}) $, on every element in a sequence,
where $h_t$ is the hidden state $t$ the time step, and $W$ the network weights. We can see that the hidden state at each time step depends on the previous hidden states, thus the order of elements (words) is important. This process also enables RNNs to handle inputs of variable length. 

RNNs are difficult to train \cite{pascanu2013a}, because gradients may grow or decay exponentially over long sequences \cite{bengio1994,hochreiter2001}. 
A way to overcome these problems is to use more sophisticated variants of regular RNNs, like Long Short-Term Memory (LSTM) networks~\cite{hochreiter1997} or Gated Recurrent Units (GRU)~\cite{cho2014a}, introducing a gating mechanism to ensure proper gradient flow through the network. 

\subsection{Self-Attention Mechanism}\label{sec:self-att}
\begin{figure}[!t]
	\captionsetup{farskip=0pt} 
	\centering
	\subfloat[Regular RNN ]{\includegraphics[width=0.48\columnwidth,page=1]{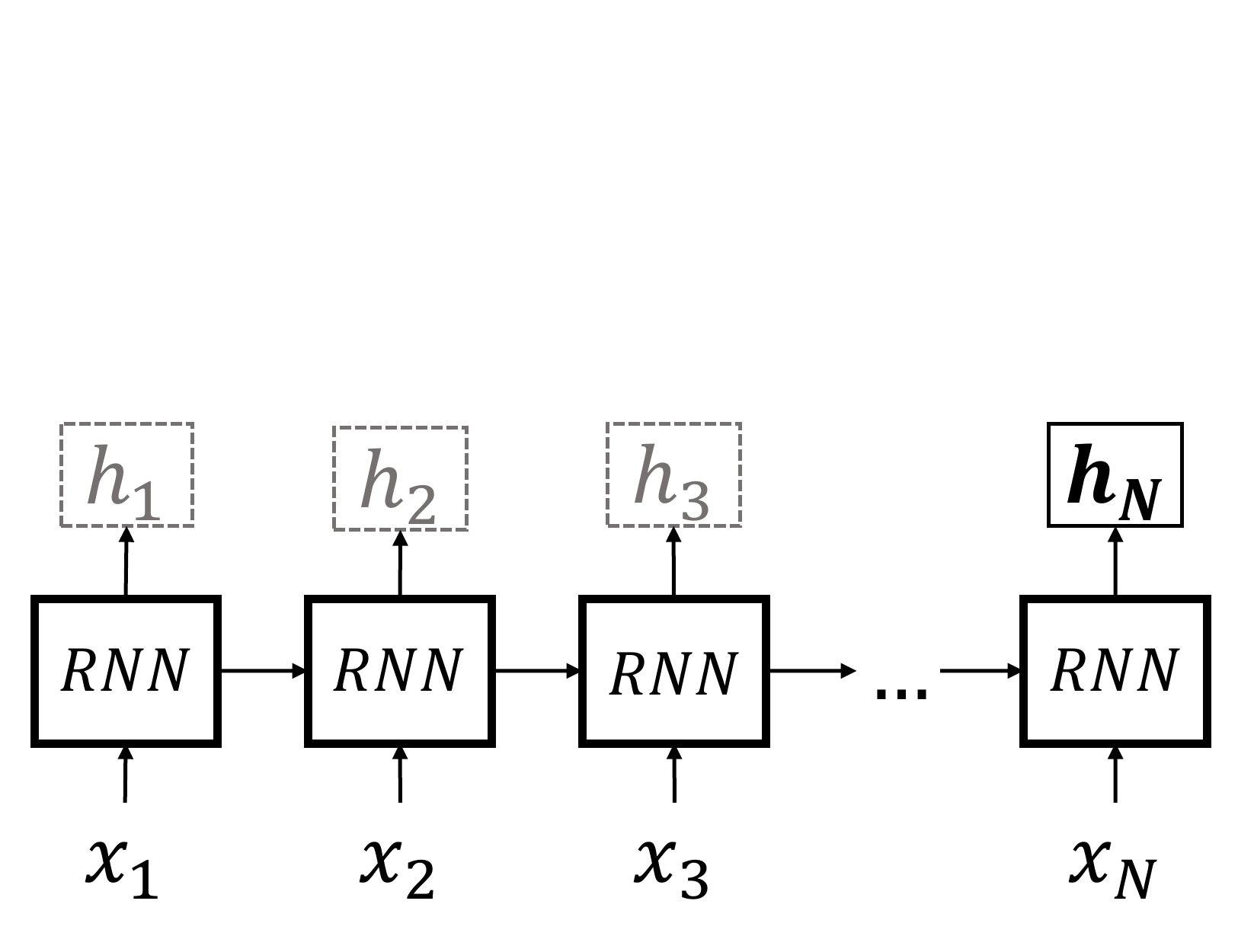}\label{fig:rnn}}
	\hfill
	\subfloat[Attention RNN]{\includegraphics[width=0.48\columnwidth,page=2]{rnn}\label{fig:rnn_att}}
	\caption{Comparison between regular RNN and attentive RNN.}
	\label{fig:attention}
\end{figure}

RNNs update their hidden state $h_i$ as they process a sequence and the final hidden state holds a summary of the information in the sequence. 
In order to amplify the contribution of important words in the final representation, a self-attention mechanism \cite{DBLP:journals/corr/BahdanauCB14} is used as shown in Fig.~\ref{fig:attention}. 
By employing an attention mechanism, the representation of the input sequence $r$ is no longer limited to just the final state $h_N$, but rather it is a combination of all the hidden states $h_i$. This is done by computing the sequence representation, as the convex combination of all $h_i$. The weights $a_i$ are learned by the network and their magnitude signifies the importance of each $h_i$ in the final representation. Formally: 
\begin{align*}
r = \sum_{i=1}^{N} a_i h_i \quad where \quad \sum_{i=1}^{N} a_i = 1, \quad a_i > 0
\end{align*}

\section{Model Description}
Next, we present in detail the submitted models. 
For all subtasks, we adopted a transfer learning approach, by pretraining a BiLSTM network with a deep attention mechanism on SA2017 dataset. Afterwards, we replaced the last layer of the pretrained model with a task-specific layer and fine-tuned the whole network for each subtask. 
\subsection{Transfer Learning Model (TF)}\label{sec:nn1} 
Our transfer learning model is based on the sentiment analysis model in~\cite{baziotis2017datastories}. It consists of a 2-layer bidirectional LSTM (BiLSTM) with a deep self-attention mechanism.
\begin{figure*}[!ht]
	\captionsetup{farskip=0pt} 
	\centering
	\includegraphics[width=0.9\textwidth]{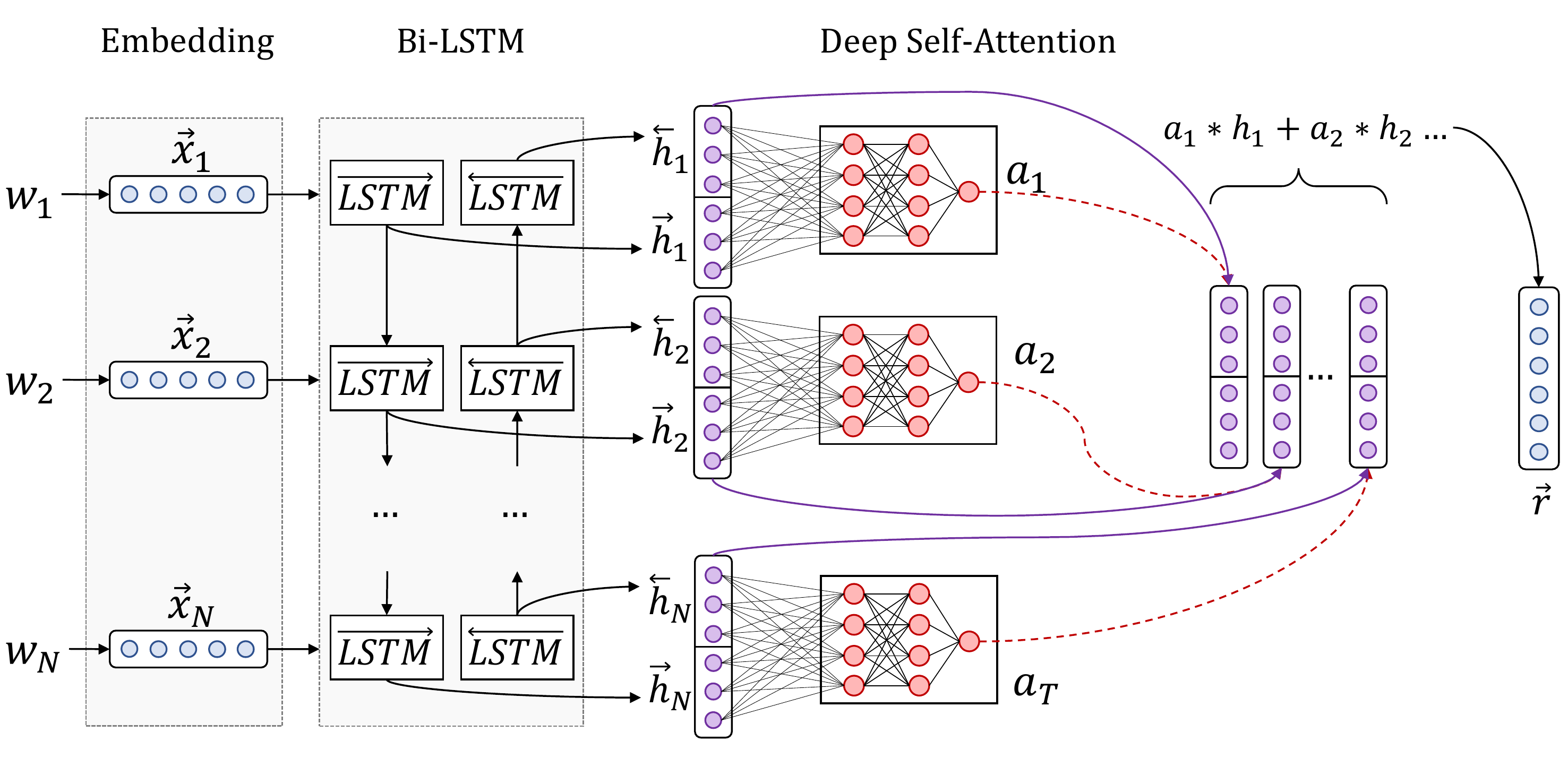}
	\caption{The proposed model, composed of a 2-layer BiLSTM with a deep self-attention mechanism. }
	\label{fig:nn1}
\end{figure*}

\noindent\textbf{Embedding Layer}. The input to the network is a Twitter message, treated as a sequence of words. We use an embedding layer to project the words $w_1,w_2,...,w_N$ to a low-dimensional vector space $ R^W$, where $W$ is the size of the embedding layer and $N$ the number of words in a tweet. We initialize the weights of the embedding layer with our pre-trained word embeddings (Section~\ref{sec:embeddings}).

\noindent\textbf{BiLSTM Layer}. An LSTM takes as input a sequence of word embeddings and produces word annotations $h_1,h_2,...,h_N$, where $h_i $ is the hidden state of the LSTM at time-step $i$, summarizing  all the information of the sentence up to $w_i$. 
We use bidirectional LSTMs (BiLSTM) in order to get word annotations that summarize the information from both directions. A BiLSTM consists of $2$ LSTMs, a forward LSTM $ \overrightarrow{f} $  that parses the sentence from $w_1$ to $w_N$ and a backward LSTM $ \overleftarrow{f} $ that parses the sentence from $w_N$ to $w_1$. We obtain the final annotation for each word $h_i$, by concatenating the annotations from both directions,
\begin{equation}
h_i = \overrightarrow{h_i} \parallel \overleftarrow{h_i}, \quad h_i \in R^{2L}
\end{equation}
where $ \parallel $ denotes the concatenation operation and $L$ the size of each LSTM. 

\noindent\textbf{Attention Layer}. 
To amplify the contribution of the most informative words, we augment our BiLSTM with a self-attention mechanism. 
We use a deep self-attention mechanism \cite{pavlopoulos2017deep}, to obtain a more accurate estimation of the importance of each word. 
The attention weight in the simple self-attention mechanism, is replaced with a multilayer perceptron (MLP), composed of $l$ layers with a non-linear activation function ($tanh$). The MLP learns the attention function $g$. The attention weights $a_i$ are then computed as a probability distribution over the hidden states $h_i$. The final representation $r$ is the convex combination of $h_i$ with weights $a_i$.
\begin{align}
e_i &= g(h_i)\label{eq:att_ei}\\
a_i &= \dfrac{exp(e_i)}{\sum_{t=1}^{N} exp(e_t)}\label{eq:att_ai}\\
r &= \sum_{i=1}^{N} a_i h_i \label{eq:att_r}, \quad r \in R^{2L}
\end{align}

\noindent\textbf{Output Layer}. We use vector $r$ as the feature representation, which we feed to a final task-specific layer. For the regression tasks, we use a fully-connected layer with one neuron and a sigmoid activation function. For the ordinal classification tasks, we use a fully-connected layer, followed by a $softmax$ operation, which outputs a probability distribution over the classes. Finally, for the multilabel classification task, we use a fully-connected layer with 11 neurons (number of labels) and a sigmoid activation function, performing binary classification for each label.

\subsection{Fine-Tuning}\label{sec:transfer}
After training a network on the pretraining dataset (SA2017), we fine-tune it on each subtask, by replacing its final layer with a task-specific layer. We experimented with two fine-tuning schemes. 
The first approach is to fine-tune the whole network, that is, both the pretrained encoder (BiLSTM) and the task-specific layer. 
The second approach is to use the pretrained model only for weight initialization, freeze its weights during training and just fine-tune the final layer. 
Based on the experimental results, the first approach obtains significantly better results in all tasks.

\subsection{Regularization}\label{sec:reg}
In both models, we add Gaussian noise to the embedding layer, which can be interpreted as a random data augmentation technique, that makes models more robust to overfitting.
In addition to that, we use dropout \cite{srivastava2014} and we stop training after the validation loss has stopped decreasing (early-stopping).

Furthermore, we do not fine-tune the embedding layers. Words occurring in the training set, are projected in the embedding space and the classifier correlates certain regions of the embedding space to certain emotions. However, words included only in the test set, remain at their initial position which may no longer reflect their ``true'' emotion, leading to mis-classifications.

\section{Experiments and Results}

\subsection{Experimental Setup}\label{sec:setup}
\noindent\textbf{Training}\label{sec:train}
We use Adam algorithm~\cite{kingma2014} for optimizing our networks, with mini-batches of size 32 and we clip the norm of the gradients~\cite{pascanu2013a} at 1, as an extra safety measure against exploding gradients. For developing our models we used PyTorch \cite{paszke2017automatic} and Scikit-learn \cite{pedregosa2011}.

\noindent\textbf{Class Weights}.
In subtasks \textit{EI-oc} and \textit{V-oc}, some classes have more training examples than others, introducing bias in our models. To deal with this problem, we apply class weights to the loss function, penalizing more the misclassification of under-represented classes. These weights are computed as the inverse frequencies of the classes in the training set.

\noindent\textbf{Hyper-parameters}.
In order to tune the hyper-parameter of our model, we adopt a Bayesian optimization \cite{bergstra2013} approach, performing a more time-efficient search in the high dimensional space of all the possible values, compared to grid or random search. 
We set size of the embedding layer to 310 (300 \textit{word2vec} + 10 affective dimensions), which we regularize by adding Gaussian noise with $\sigma = 0.2$ and dropout of 0.1. The sentence encoder is composed of 2 BiLSTM layers, each of size 250 (per direction) with a 2-layer self-attention mechanism. Finally, we apply dropout of 0.3 to the encoded representation.

\begingroup
\setlength{\tabcolsep}{3pt} 
\renewcommand{\arraystretch}{1.1} 
\begin{table*}[!t]
	\centering
	\small
	\begin{tabular}{|l|r|r|r|r|r|r|r|r|r|r|r|}
		\hline
		& \multicolumn{4}{c|}{\textbf{EI-reg (pearson)}}                         & \multicolumn{4}{c|}{\textbf{EI-oc (pearson)}}                          & \multicolumn{1}{c|}{\multirow{2}{*}{\textbf{\begin{tabular}[c]{@{}c@{}}V-Reg \\ (pearson)\end{tabular}}}} & \multicolumn{1}{c|}{\multirow{2}{*}{\textbf{\begin{tabular}[c]{@{}c@{}}V-oc\\ (pearson)\end{tabular}}}} & \multicolumn{1}{c|}{\multirow{2}{*}{\textbf{\begin{tabular}[c]{@{}c@{}}E-c \\ (jaccard)\end{tabular}}}} \\ \cline{1-9}
		                     & \textbf{anger}  & \textbf{fear}   & \textbf{joy}    & \textbf{sadness} & \textbf{anger}  & \textbf{fear}   & \textbf{joy}    & \textbf{sadness} & \multicolumn{1}{c|}{} & \multicolumn{1}{c|}{} & \multicolumn{1}{c|}{} \\ \hline
		\textbf{BOW}         & 0.5249          & 0.5227          & 0.5716          & 0.4721           & 0.3996          & 0.3491          & 0.4456          & 0.3835           & 0.5963                & 0.4954                & 0.4572                \\ \hline
		\textbf{NBOW}   	 & 0.6539          & 0.6318          & 0.6355          & 0.6305           & 0.5573          & 0.3796          & 0.5044          & 0.5009           & 0.7501                & 0.6527                & 0.4541                \\ \hline
		\textbf{NBOW+A*}  	 & 0.656           & 0.6359          & 0.6384          & 0.6341           & 0.5367          & 0.3906          & 0.4803          & 0.5005           & 0.7457                & 0.6578                & 0.4478                \\ \hline
        \hhline{|=|=|=|=|=|=|=|=|=|=|=|=|}
		\textbf{LSTM-RD}     & 0.7568          & \textbf{0.7357} & 0.7313          & 0.7479           & \textbf{0.6387} & \textbf{0.5874} & 0.6226          & 0.6343           & \textbf{0.8462}       & \textbf{0.7722}       & \textbf{0.5788}       \\ \hline
		\textbf{LSTM-TL-FR}  & 0.7347          & 0.6509          & 0.7321          & 0.7269           & 0.5999          & 0.4666          & 0.6264          & 0.6030           & 0.8275                & 0.7331                & 0.5243                \\ \hline
		\textbf{LSTM-TL-FT}  & \textbf{0.7717} & 0.7273          & \textbf{0.7638} & \textbf{0.7665}  & 0.6329          & 0.5702          & \textbf{0.6351} & \textbf{0.6400}  & 0.8390                & 0.7652                & \textbf{0.5788}       \\ \hline
	\end{tabular}
	\caption{Results of our experiments across all subtasks on the official evaluation metrics. For subtasks \textit{EI-reg, EI-oc, V-reg, V-oc}, the evaluation metric is Pearson correlation. For subtask \textit{E-c}, the evaluation metric is multi-label accuracy (Jaccard index). 
    BOW stands for Bag-of-Words baseline, N-BOW stands for Neural Bag-of-Words baseline and N-BOW+A indicates the inclusion of the affective word features.
    As for the neural models, RD stands for random initialization, TL for Transfer Learning, FR for Frozen pretrained layers (without fine-tuning) and FT for Fine-Tuning. For our deep-learning models, the results are computed by averaging $10$ runs to account for the variability in training performance.
    }
	\label{table:comp}
\end{table*}


\endgroup

\subsection{Experiments}\label{sec:experiments}

In Table~\ref{table:comp}, we compare the proposed transfer learning models against 3 strong baselines. Pearson correlation is the metric used for the first four subtasks, whereas Jaccard index is used for the \textit{E-c} multi-label classification subtask. The first baseline is a unigram Bag-of-Words (BOW) model with TF-IDF weighting. The second baseline is a Neural Bag-of-Words (N-BOW) model, where we retrieve the \textit{word2vec} embeddings of the words in a tweet and compute the tweet representation as the average (centroid) of the constituent \textit{word2vec} embeddings. Finally, the third baseline is similar to the second one, but with the addition of 10-dimensional affective embeddings that model affect-related dimensions (valence, dominance, arousal, etc). 
Both BOW and N-BOW features are then fed to a linear SVM classifier, with tuned $C=0.6$. 
In order to assess the impact of transfer learning, we evaluate the performance of each model in 3 different settings:   
\begin{enumerate*}[(1)]
	\item random weight initialization (LSTM-RD),
	\item transfer learning with frozen weights (LSTM-TL-FR),
	\item transfer learning with finetuning (LSTM-TL-FT).
\end{enumerate*}
The results of our neural models in Table~\ref{table:comp} are computed by averaging the results of $10$ runs to account for model variability. 

\noindent\textbf{Baselines.} 
Our first observation is that N-BOW baselines significantly outperform BOW in subtasks \textit{EI-reg}, \textit{EI-oc}, \textit{V-reg} and \textit{V-oc}, in which we have to predict the intensity of an emotion, or the tweet's valence. However, BOW achieves slightly better performance in subtask \textit{E-c}, in which we have to recognize the emotions expressed in each tweet.
This can be attributed to the fact that BOW models perform well in tasks where we the occurrence of certain words is sufficient, to accurately determine the classification result. This suggests that in subtask E-c, certain words are highly indicative of some emotions. 
Word embeddings, though, that encode the correlation of each word with different dimensions, enable NBOW to better predict the intensity of various emotions.
Furthermore, regarding the affective embeddings, we can directly observe their impact by the performance gain over the NBOW baseline.

\noindent\textbf{Transfer Learning.} 
We observe that our neural models achieved better performance than all baselines by a large margin.
Moreover, we can see that our transfer learning model yielded higher performance over the non-transfer model in most of the Emotion Intensity (EI) subtasks. In the Emotion multi-label classification subtask (E-c), transfer learning did not outperform the random initialization model. This can be attributed to the fact that our source dataset (SA17) was not diverse enough to boost the model performance when classifying the tweets into none, one or more of a set of 11 emotions. 
As for fine-tuning or freezing the pretrained layers, the overall results show that enabling the model to fine-tune always results in significant gains. This is consistent with our intuition that allowing the weights of the model to adapt to the target dataset, thus encoding task-specific information, results in performance gains. 
Regarding the emotion of \textit{joy}, we observe that in EI-reg and EI-oc subtasks, LSTM-RD matches the performance of LSTM-TL-FR. We interpret this result as an indication of the semantic similarity between the source and the target task.

\begin{table}[!bt]
	\captionsetup{farskip=0pt} 
	\centering
	\small
	\begin{tabular}{|l|r|r|r|}
		\hline
		                 & \textbf{Ave.diff. Overall} & \textbf{Ave.diff.} & \textbf{p-value} \\ \hline
		\textbf{Anger}   & 0.001                      & 0                  & 0.02223          \\ \hline
		\textbf{Fear}    & -0.003                     & -0.003             & 0                \\ \hline
		\textbf{Joy}     & 0.004                      & 0.010              & 0                \\ \hline
		\textbf{Sadness} & 0.002                      & -0.002             & 0                \\ \hline
		\textbf{Valence} & 0.005                      & 0.005              & 0                \\ \hline
	\end{tabular}
	\caption{Analysis for inappropriate biases}\label{t:mystery}
\end{table}

\noindent\textbf{Mystery dataset}. The submitted models were also evaluated against a mystery dataset, in order to investigate if there is statistically significant social bias in them. This is a very important experiment, especially when automated machine learning algorithms are interacting with social media content and users in the wild.
The mystery dataset consists of pairs of sentences that differ only in the social context (e.g. gender or race). Submitted models are expected to predict the same affective values for both sentences in the pair. The evaluation metric is the average difference in prediction scores per class, along with the p-value score indicating if the difference is statistically significant. Results are summarized in Table~\ref{t:mystery}.

\subsection{Attention visualizations}\label{sec:attentions}
\begin{figure*}[!t]

    \centering
    
	\begin{minipage}{.45\textwidth}
        \begin{mdframed}
            \hspace{20pt}\includegraphics[scale=0.7,page=15]{int_new}
            \label{fig:joy1}
            \vskip 5pt
            \hspace{20pt}\includegraphics[scale=0.7,page=10]{int_new}
        \end{mdframed}
        \vskip -10pt
		\caption{Examples of intensity of \textit{joy}}
		\label{fig:joy2}
        
        \vskip 5pt

		\begin{mdframed}
          \hspace{20pt}\includegraphics[scale=0.7,page=3
          ]{int_new}
          \label{fig:sadness1}
          \vskip 5pt
          \hspace{20pt}\includegraphics[scale=0.7,page=18]{int_new}
        \end{mdframed}
        \vskip -10pt
		\caption{Examples of intensity of \textit{sadness}}
		\label{fig:sadness}
              
        \vskip 5pt
        
        \begin{mdframed}
          \hspace{20pt}\includegraphics[scale=0.7,page=4]{int_new}
          \vskip 5pt
          \hspace{20pt}\includegraphics[scale=0.7,page=7]{int_new}
        \end{mdframed}
        \vskip -10pt
		\caption{Examples of intensity of \textit{fear}}
		\label{fig:fear}
              
        \vskip 5pt
        
        \begin{mdframed}
            \hspace{20pt}\includegraphics[scale=0.7,page=1]{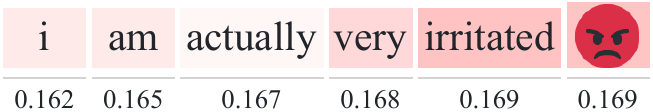}
            \vskip 5pt
            \hspace{20pt}\includegraphics[scale=0.7,page=2]{int_new}
        \end{mdframed}
        \vskip -10pt
		\caption{Examples of intensity of \textit{anger}}
		\label{fig:anger}
	\end{minipage}%
    \hfill
	\begin{minipage}{0.45\textwidth}
      \begin{mdframed}
        \hspace{15pt}\includegraphics[scale=0.5]{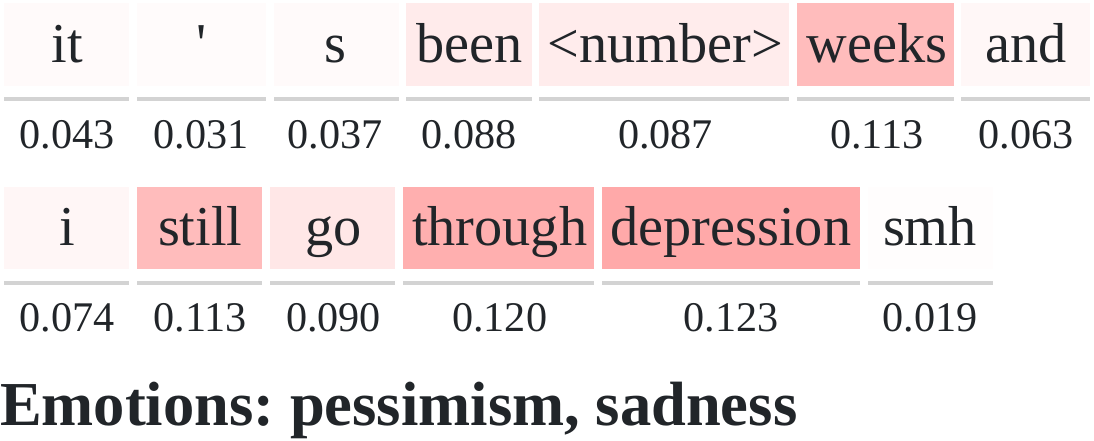}
        \label{fig:emo1}
        \vskip 19pt
        \hspace{15pt}\includegraphics[scale=0.5]{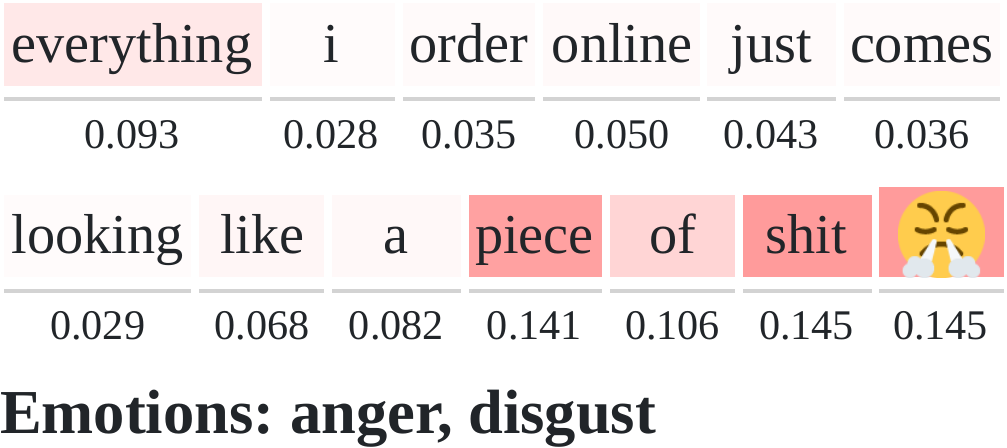}
        \label{fig:emo2}
        \vskip 19pt
        \hspace{15pt}\includegraphics[scale=0.5]{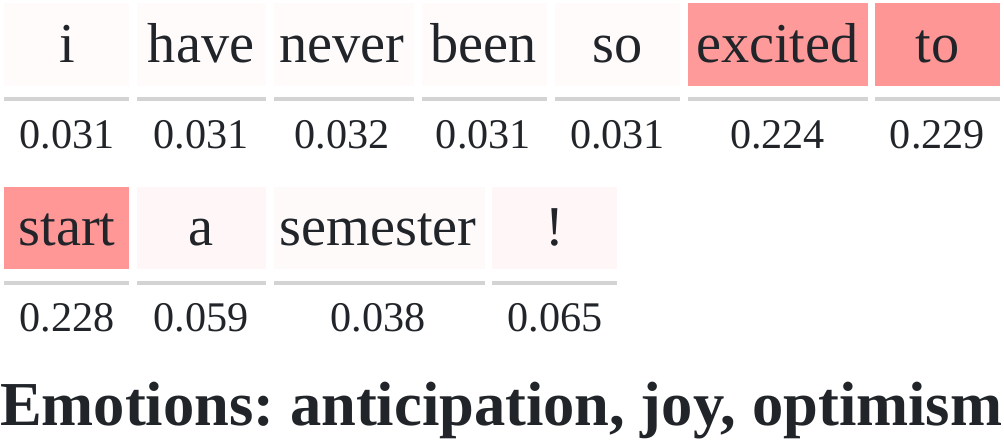}
        \label{fig:emo3}
        \vskip 17pt
        \hspace{15pt}\includegraphics[scale=0.5]{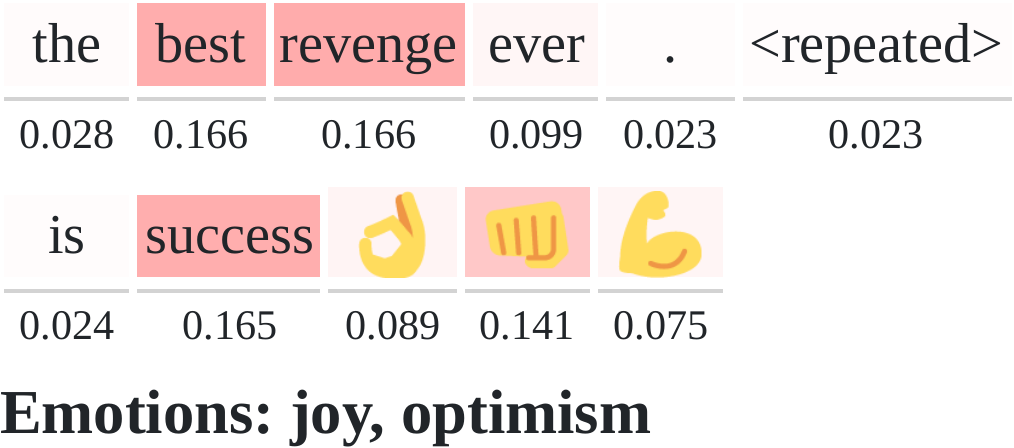}
        \label{fig:emo4}
      \end{mdframed}
      \vskip -10pt
      \caption{Examples of emotion recognition}
	\end{minipage}

\caption{Attention heat-map visualization. 
The color intensity of each word corresponds to its weight (importance), given by the self-attention mechanism (Section \ref{sec:self-att}).}
\label{fig:attention-heatmaps}

\end{figure*}

Fig.~\ref{fig:attention-heatmaps} shows a heat-map of the attention weights on top of 8 example tweets (2 tweets per emotion). The color intensity corresponds to the weight given to each word by the self-attention mechanism and signifies the importance of this word for the final prediction. We can see that the salient words correspond to the predicted emotion (e.g. ``irritated'' for anger, ``mourn'' for sadness etc.). An interesting observation is that when emojis are present they are almost always selected as important, which indicates their function as weak annotations. Also note that the attention mechanism can hint to dependencies between words even if they far in a sentence, like the ``why'' and ``mad'' in the sadness example.


\subsection{Competition Results}
Our official ranking was 2/48 in subtask 1A (EI-reg), 5/39 in subtask 2A (EI-oc), 4/38 in subtask 3A (V-reg), 8/37 (tie with 6 and 7 place) in subtask 4A (V-oc) and 1/35 in subtask 5A (E-c). All of our models achieved competitive results. We used the same transfer learning approach in  all subtasks (LSTM-TL-FT), utilizing the same pretrained model.

\section{Conclusion}
In this paper we present a deep-learning system for short text emotion intensity, valence estimation for both regression and classification and multi-class emotion classification. We used Bidirectional LSTMs, with a deep attention mechanism and took advantage of transfer learning in order to address the problem of limited training data. 

Our models achieved excellent results in single and multi-label classification tasks, but mixed results in emotion and valence intensity tasks. Future work can follow two directions. Firstly, we aim to revisit the task with different transfer learning approaches, such as \cite{felbo2017using,howard2018fine,hashimoto2016joint}. Secondly, we would like to introduce character-level information in our models, based on \cite{wieting2016charagram,labeau2017character}, in order to overcome the problem of out-of-vocabulary (OOV) words and learn syntactic and stylistic features \cite{peters2018deep}, which are highly indicative of emotions and their intensity.

Finally, we make both our pretrained word embeddings and the source code of our models available to the community\footnote{\url{github.com/cbaziotis/ntua-slp-semeval2018-task1}}, 
in order to make our results easily reproducible and facilitate further experimentation in the field.
\newline

\noindent\textbf{Acknowledgements}. 
This work has been partially supported by the BabyRobot project supported by EU H2020 (grant \#687831). Also, the authors would like to thank NVIDIA for supporting this work by donating a TitanX GPU.

\bibliography{refs}
\bibliographystyle{acl_natbib}

\appendix

\end{document}